\documentclass[runningheads]{llncs}

\usepackage{microtype}
\usepackage{soul}
\usepackage{url}
\usepackage{graphicx}
\usepackage[utf8]{inputenc}
\usepackage{booktabs} 
\usepackage{vwcol}
\usepackage{hyperref}
\usepackage[linesnumbered,ruled,vlined]{algorithm2e}
\usepackage{caption}
\usepackage{subcaption}
\usepackage{amsmath}
\usepackage{amssymb}
\usepackage{paralist}
\usepackage[title]{appendix}
\def\bx{{\bf x}}

\def\bu{{\bf u}}

\def\bw{{\bf w}}

\newcommand{\eat}[1]{}
\begin{document}
%
\title{On Universalized Adversarial and Invariant Perturbations}
%
%
\author{Sandesh Kamath\inst{1} \and
Amit Deshpande\inst{2} \and
K V Subrahmanyam\inst{1}}
%
%
\institute{Chennai Mathematical Institute, Chennai, India \and
Microsoft Research, Bengaluru, India\\
\email{{ksandeshk,kv}@cmi.ac.in}\\
\email{amitdesh@microsoft.com}}
%
\maketitle              
\begin{abstract}
Convolutional neural networks or standard CNNs (StdCNNs) are translation-equivariant models that achieve translation invariance when trained on data augmented with sufficient translations. Recent work on equivariant models for a given group of transformations (e.g., rotations) has lead to group-equivariant convolutional neural networks (GCNNs). GCNNs trained on data augmented with sufficient rotations achieve rotation invariance. Recent work by authors \cite{S20invariance} studies a trade-off between invariance and robustness to adversarial attacks. In another related work \cite{S20universal}, given any model and any input-dependent attack that satisfies a certain spectral property, the authors propose a universalization technique called SVD-Universal to produce a universal adversarial perturbation by looking at very few test examples. In this paper, we study the effectiveness of SVD-Universal on GCNNs as they gain rotation invariance through higher degree of training augmentation. We empirically observe that as GCNNs gain rotation invariance through training augmented with larger rotations, the fooling rate of SVD-Universal gets better. To understand this phenomenon, we introduce universal invariant directions and study their relation to the universal adversarial direction produced by SVD-Universal.
\end{abstract}

\section{Introduction}

Universal adversarial perturbations are input-agnostic so that the same perturbation fools the trained model on a large fraction of test inputs. Recent work by \cite{Dezfooli17,Dezfooli17anal} on universal adversarial perturbations looks at the curvature of the decision boundary. Their universal attacks are more sophisticated than the simple and fast, gradient-based adversarial perturbations, and require significantly more computation. Dezfooli et al. \cite{Dezfooli17anal} give a theoretical justification for existence of universal adversarial perturbations using directions in which the decision boundary has positive curvature. 
Khrulkov and Oseledets \cite{Khrulkov18} propose a different method to construct a universal adversarial perturbation. Their perturbation is based on computing the $(p,q)$-singular vectors of the Jacobian matrices of hidden layers of the network. The authors then judiciously select a particular layer to construct an attack and for that layer they choose the best possible $(p,q)$-singular attack vector. 
In \cite{S20universal}, the authors propose a simpler technique to obtain universal adversarial perturbations from input-dependent perturbations such as Gradient, FGSM or DeepFool. 
The authors show that the following universalization technique yields universal adversarial perturbations whose fooling rates are comparable to \cite{Dezfooli17} but weaker than that obtained in \cite{Khrulkov18}. {\it Take a small random sample of size 64 of the test data and for each point in the sample, construct an input dependent adversarial attack vector. Output the top singular vector of these adversarial attack vectors as the universal adversarial perturbation.} Unlike in \cite{Khrulkov18} where the authors need access to the various layers of the neural network, and where the authors compute  
an approximation to the $(\infty, 10)$-singular vector, the computational overheads in \cite{S20universal} are significantly less. Furthermore, the authors in \cite{S20universal} support their technique with 
theoretical bounds.

Geometric transformations such as rotations, translations are simple input transformations that are model-agnostic and input-agnostic and work as natural attacks during test time, see \cite{Dumont18}, \cite{Gilmer18}. One way to counter such attacks is to use neural network models that are translation and rotation-equivariant by construction. Standard Convolutional Neural Networks (StdCNNs) are translation-equivariant but not equivariant with respect to other spatial symmetries such as rotations, reflections etc. Variants of CNNs to achieve rotation-equivariance and other symmetries have received much attention recently, notably, Harmonic Networks (H-Nets) \cite{Worrall16}, cyclic slicing and pooling \cite{Dieleman16}, Tranformation-Invariant Pooling (TI-Pooling) \cite{Laptev16}, Group-equivariant Convolutional Neural Networks (GCNNs) \cite{Cohen16}, Steerable CNNs \cite{Cohen17}, Deep Rotation Equivariant Networks (DREN) \cite{Li17}, Rotation Equivariant Vector Field Networks (RotEqNet) \cite{Marcos17}, Polar Transformer Networks (PTN) \cite{Esteves18}. However, we use GCNNs as the representative rotation equivariant networks since they have both, good invariance properties, and a theoretical justification based on representation theory \cite{Kondor18}. 

A natural question is to study the fooling rates of the universal adversarial perturbations from \cite{Dezfooli17anal}, \cite{Khrulkov18} and  \cite{S20universal} on  equivariant networks such as the above. 
In this paper, we extend the empirical analysis of \cite{S20universal} to rotation equivariant networks with augmentation. Since these networks achieve invariance to rotations with augmentation, a natural question is whether there are perturbations to which a well trained model is agnostic, so even a large perturbation along those directions does not fool the network? We show that this is indeed the case and call such directions, universal invariant perturbations. One would expect these directions to be orthogonal to the universal adversarial perturbations. We empirically show that this is the case. 

To address these questions we use StdCNNs which are translation equivariant networks. For the rotation-equivariant networks we use GCNNs as a representative network. However, both StdCNNs and GCNNs require rotation augmentation at training time to be invariant to rotations. GCNNs based on steerable filters, have a solid theoretical justification as shown by Kondor and Trivedi \cite{Kondor18}, and achieve nearly state of the art results on MNIST-rot and CIFAR-10 as reported by Esteves et al. \cite{Esteves18}. In the main paper we only report experiments on the dataset CIFAR-10. Experiments on MNIST are reported in the Appendix \ref{svd:mnist}. 

\subsection{Connection between fooling rate and error rate}
In some of our experiments we compute the fooling rate of perturbations, and in some others we compute the error rate of perturbations. We show in Appendix~\ref{sec:defn} that the difference between the two is marginal for networks trained to achieve high accuracy.  The proof is borrowed from  \cite{S20universal}[Appendix A].

\section{SVD-Universal for models trained with larger rotation augmentations}
There has been a lot of work on making CNN's robust to geometric transformations like rotations and translations of input images so that the label assigned by a classifier to an image and its rotated/translated version is the same. GCNNs~\cite{Cohen16}, Harmonic nets~\cite{Worrall16} and Clebsch Gordon networks \cite{Kondor18Clebsch} are some neural networks designed to be  equivariant to such geometric transformations. To make these networks equivariant to such transformation it is necessary to augment the train data. So in addition to upright images, rotations of images are also added to the training phase. In this section we evaluate the error rate of SVD-Universal attacks on StdCNNs and GCNNs trained with data augmented with rotations. To set the stage, in Section \ref{subsec:gcnn:cifar10}, we first give the error rate of these attacks on GCNNs when the training data and test data are {\it not augmented} with rotations. 

\subsection{Error rates for GCNN's on CIFAR-10} \label{subsec:gcnn:cifar10}
 For a fair comparison with the results in ~\cite{S20universal}, the GCNN configuration used for CIFAR-10 is the same as that of ResNet18, except that the operations going from one layer to the next are replaced by equivalent GCNN operations,  as given in \cite{Cohen16}.  Similarly the GCNN used for MNIST, has the same configuration as the StdCNN configuration given in Table~\ref{gcnn-table} with equivalent GCNN operations as given in \cite{Cohen16}. 
 
\begin{figure}[!h]
\begin{center}
\includegraphics[width=0.49\linewidth]{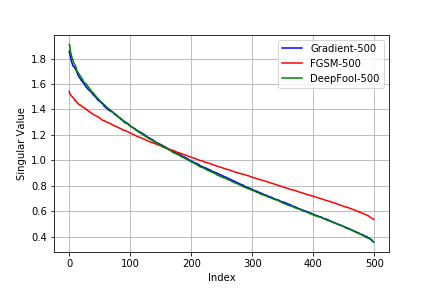} 
\includegraphics[width=0.49\linewidth]{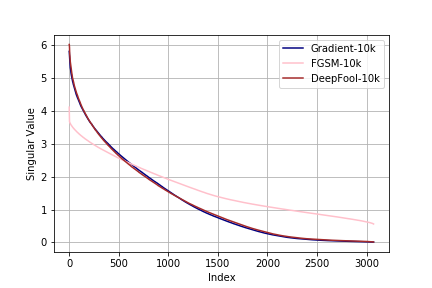} 
\caption{On CIFAR-10, GCNN/ResNet18, Singular values of attack directions over a sample of (left) 500 and (right) 10,000 test points.}
\label{fig:cifar10-gcnn-dot-prod}
\end{center}
\end{figure}

\begin{figure}[h!]
\begin{center}
\includegraphics[width=0.49\linewidth]{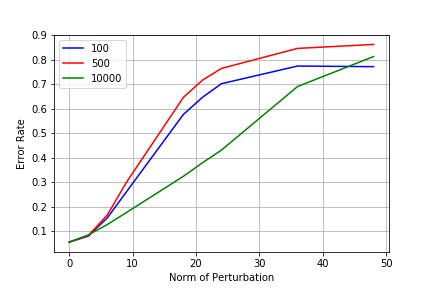}
\includegraphics[width=0.49\linewidth]{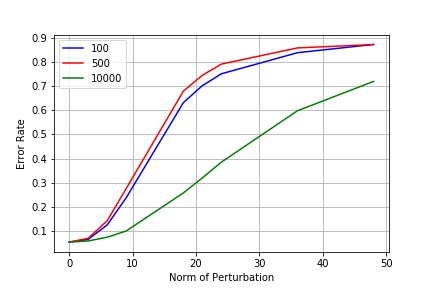}
\includegraphics[width=0.49\linewidth]{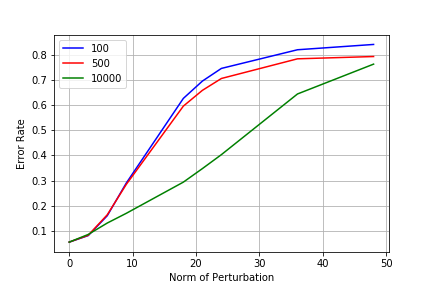}
\end{center}
\caption{On CIFAR-10, GCNN: error rate vs. norm of perturbation along top singular vector of attack directions on 100/500/10000 samples, (top left) Gradient (top right) FGSM (bottom) DeepFool}
\label{fig:gcnn-fool}
\end{figure}

Recall that the theoretical justification for the existence of universal adversarial perturbations in ~\cite{S20universal}, is the drop in singular values of the matrix of attack directions for a small sample of test inputs of CIFAR-10. We show that the same phenomenon holds for GCNNs also. We first trained the GCNN to achieve high accuracy on CIFAR-10. We take a sample of test inputs of CIFAR-10 and using the trained GCNN, obtain input dependent adversarial attack vectors (Gradient, FGSM and DeepFool attacks).  We plot the singular values of the matrix of attack vectors in Figure \ref{fig:cifar10-gcnn-dot-prod}, and observe that the graph is similar to that in \cite{S20universal}[Figure 3] - the drop in singular values is a shared phenomenon across all the input dependent attacks, exactly as reported in  ~\cite{S20universal} for StdCNNs. We then used the top singular vector of these input dependent adversarial attack vectors as a universal perturbation, SVD-Universal.  We plot the error rates for SVD-Universal obtained with various sample sizes in Figure \ref{fig:gcnn-fool}. When the $\ell_2$ norm of the universal perturbation constructed from DeepFool  (Figure  \ref{fig:gcnn-fool}., bottom) using a sample of size 100 is scaled to 8, the error rate is more than 25\%. Since the error rate of the trained GCNN on CIFAR-10 is about 5\%,  it follows from the results in Appendix~\ref{sec:defn}, that this universal adversarial perturbation fools more than 20\% of the test inputs.

\subsection{SVD-Universal for models trained with larger rotation augmentations} \label{sub:large-rot}
\begin{figure*}[!h]
\begin{center}
\includegraphics[width=0.49\linewidth]{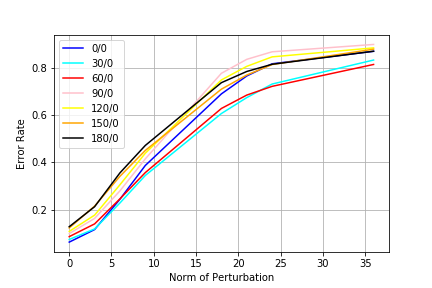}
\includegraphics[width=0.49\linewidth]{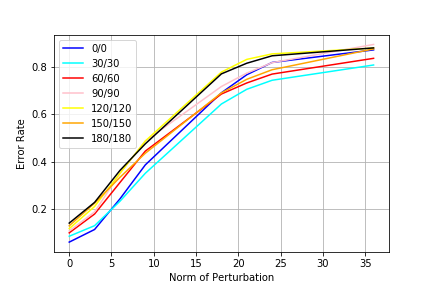}
\end{center}
\caption{On CIFAR-10, error rate of our universal attack using top singular vector of 100 test point gradients, for StdCNN/ResNet18 train-augmented with random rotations in range $[-\theta^{\circ}, \theta^{\circ}]$ (left) test unrotated, (right) test-augmented with the same range of rotations as the training set. }
\label{fig:cifar10-stdcnn-100-grad-unrot-rot-fool}
\end{figure*}

\begin{figure*}[h!]
\begin{center}
\includegraphics[width=0.49\linewidth]{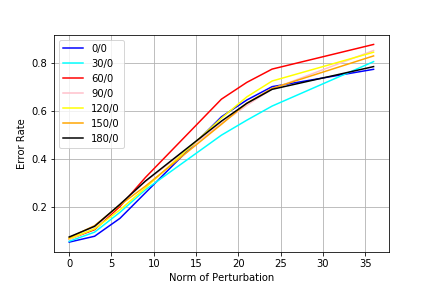}
\includegraphics[width=0.49\linewidth]{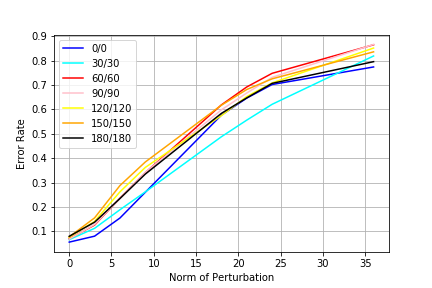}
\end{center}
\caption{On CIFAR-10, error rate of our universal attack using top singular vector of 100 test point gradients, for GCNN/ResNet18 train-augmented with random rotations in range $[-\theta^{\circ}, \theta^{\circ}]$ (left) test unrotated, (right) test-augmented with the same range of rotations as the training set. }
\label{fig:cifar10-gcnn-100-grad-unrot-rot-fool}
\end{figure*}

We also evaluate the error rate of SVD-Universal attacks on equivariant networks when the training data is augmented with rotations. We report this on the CIFAR-10 data set in \eat{Figures~\ref{fig:mnist-gcnn-100-grad-unrot-rot-fool}}Figures~\ref{fig:cifar10-stdcnn-100-grad-unrot-rot-fool} and ~\ref{fig:cifar10-gcnn-100-grad-unrot-rot-fool}. The equivariant network architecture is discussed in section~\ref{app:arch}.

We augment the networks with random rotations of training data in the range $\pm 180^{\circ}$.  Figure~\ref{fig:cifar10-stdcnn-100-grad-unrot-rot-fool}, left, shows that SVD-Universal attack vectors improve their error rate when ResNets are trained to handle larger rotations - the error rate of  SVD-Universal perturbation of norm 10 constructed from unaugmented test data ($0^{\circ}$),  is about about 50\% on the (unaugmented) test data when the network is trained to handle rotations of $180^{\circ}$ (black line in the figure), as against an error rate of 40\% when it is not trained to handle rotations. There is no significant difference between the figures on the left and on the right in Figure 4. While 0/0 and 180/0 difference is about 10\% in StdCNNs, 0/0 and 180/0 is very different for rotation-equivariant GCNNs. \eat{For a comparison look at SVD universal vectors of norm 10. When the network is not trained to handle rotations the error rate is about 43\% (blue line) but when the network is trained to handle rotations upto 180\% the error rate is almost 55\% (black line).} For GCNNs we observe a similar trend, the universal attack shows a larger error rate on networks trained to handle larger rotations (compare the black line in Figure~\ref{fig:cifar10-gcnn-100-grad-unrot-rot-fool}, left, with the blue line in the same figure). The Figures on the right were obtained by training the network with data augmented with rotations upto $\pm180^{\circ}$. The test data is also augmented with random rotations upto $\pm180^{\circ}$. And our universal perturbation is obtained from sampling this augmented data. Both, in StdCNNs, as well as in GCNNs, the error rate remains almost the same, even as test is augmented with rotations that the model was trained with. 

  The construction of adversarial attacks that work for a given image as well as its geometric transformations (especially, rotations) has been of interest in literature, e.g., see \cite{Athalye17}.  

\paragraph{Convention used in the legends of our figures.} We use the following convention in the legends of some plots. A coloured line labeled $A/B$ indicates that the training data is augmented with random rotations from $[-A^{\circ}, A^{\circ}]$ and the test data is augmented with random rotations from $[-B^{\circ}, B^{\circ}]$. If $A$ (resp. $B$) is zero it means the training data (resp. test data) is unrotated. If the model is trained with random rotations from $[-A^{\circ}, A^{\circ}]$ and the test data is randomly rotated with varying $B$ to draw the plot, we only mention $A$ and not $B$, which is self-explanatory.

\clearpage

\section{Universal invariant perturbations}
Let $\bx$ be an image for which the predicted label of a neural network classifier $f$ is $f(\bx)$. For highly accurate classifiers trained on data that already contains small rotation augmentations, one would expect $f(\bx + \Delta(\bx)) = f(\bx)$, where $\bx + \Delta(\bx)$ is the image resulting from a small rotation applied to $\bx$. For an image $\bx$, we call $\mathcal{I}(\bx)$ an invariant perturbation for $\bx$, if $f(\bx + \mathcal{I}(\bx)) = f(\bx)$. Using this terminology we expect $\Delta(\bx)$ to be an invariant perturbation for $\bx$ for classifiers trained on data that is inherently (or additionally) augmented with small rotations. A natural question is whether there are any universal invariant perturbations $\bx + \bu$ that are invariant for most $\bx$. We show empirically that there are such universal invariant perturbations. By their very definition we would also expect that universal invariant perturbations will not be universal adversarial perturbations.  We show this empirically by plotting the error rates of the top singular invariant perturbation of CIFAR-10 on StdCNN/ResNet18 and GCNN/ResNet18 in Figure~\ref{stdcnn-gcnn-top-rvh}. The error rate is below 0.2 even when the norm of the invariant perturbation is scaled up to 50.

For illustration purposes we augment Figure 1 in \cite{S20universal} with $g'_{1}, g'_{2}, g'_{3}$ which are orthogonal to $g_{1}, g_{2}, g_{3}$ respectively, the minimal $\ell_{2}$-norm perturbations such that $f(x_{i} + g'_{i}) = f(x_{i})$. These $g'_{i}, i \in \{1,2,3\}$ are examples of input dependent invariant perturbations. Similar to $v$, a universal adversarial perturbation, we show there is a $v'$ which is universal invariant perturbation, orthogonal to $v$.

\begin{figure}[!h]
\begin{center}
\includegraphics[width=0.3\linewidth]{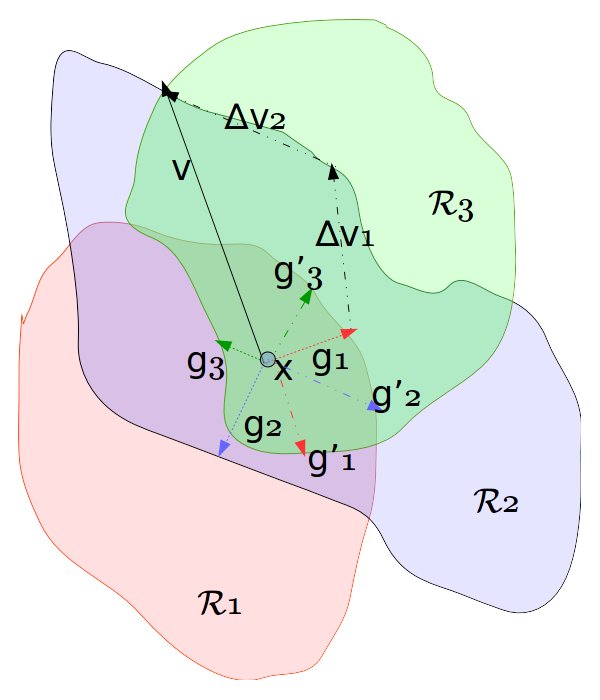}
\caption{Illustration of the universal adversarial attack problem.}
\label{fig:universal-fig}
\end{center}
\end{figure}

 We give more empirical evidence of such invariant perturbations - we show that the principal angles between the subspace of the top 5 universal adversarial perturbations and the subspace of the top 5 universal invariant perturbations are all close to $90^{\circ}$. Since universal invariant directions are easy to compute, this opens up the possibility of computing adversarial attacks   from invariant directions. We also view these invariant directions as images (refer section \ref{sub:visual-mnist}) - they seem to show some interesting structure. This could be of independent interest in the study of rotation-equivariance and steerable filters.

\begin{figure*}[!h]
\begin{center}
\includegraphics[width=0.6\linewidth]{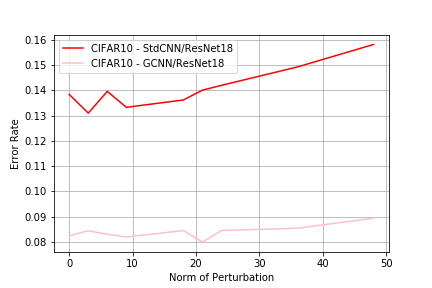}
\end{center}
\caption{Error rate vs norm of perturbation using top singular invariant vector for StdCNN and GCNN on CIFAR-10 test data.}
\label{stdcnn-gcnn-top-rvh}
\end{figure*}

\subsection{Principal components of attack and invariant directions} \label{app:invariant} 
To construct the matrix, we pick a batch of 100 images, and for each image x in our sample we compute the difference between x and its 2 degree rotation. 
 We stack these (input dependent invariant directions) as rows of the matrix and compute the top singular (unit) vector $\bw$. We call ${\bf w}$ the universal invariant perturbation. Figure~\ref{stdcnn-gcnn-top-rvh} shows the error rate of $\epsilon \bw$ as a function of $\epsilon$ on CIFAR-10 for StdCNNs and GCNNs. The curves are flat indicating that these directions are invariant perturbations. 

Table \ref{tab:principal-angles} shows deeper analysis using principal angles between subspaces. The two 5-dimensional SVD subspaces of adversarial directions and invariant directions, respectively, have nearly $90^{\circ}$ principal angles. 

Recall that for two subspaces $V$,$W$, the first principal angle is defined as the minimum angle between two unit vectors $v_1 \in V$, $w_1 \in W$. The second principal angle is the minimum angle between unit vectors $v_2 \in V, w_2 \in W$, with $v_2 \perp v_1$ and $w_2 \perp w_1$. The other principal angles are defined similarly.

{\small
\begin{table*}[!h]
\caption{Principal angles between Top-5 SVD-subspace of gradient directions of test points, and the Top-5 SVD-subspace of invariant directions on test points (small $2^{\circ}$ rotations) respectively.}
\label{tab:principal-angles}
\begin{center}
\resizebox{1.0\linewidth}{!} {
\begin{tabular} {| l | l | c | c | c | c | c |}
\hline
Dataset & Model & 1 & 2 & 3 & 4 & 5 \\
\hline
CIFAR-10 & StdCNN/ResNet18 & 89.99 & 89.83 & 89.53 & 88.92 & 88.65 \\
\hline
CIFAR-10 & GCNN/ResNet18 & 89.91 & 89.82 & 89.60 & 89.32 & 88.97 \\
\hline
\end{tabular}
}
\end{center}
\end{table*}
}



\subsection{Visualizing universal invariant perturbations} \label{sub:visual-mnist}
\begin{figure}[!]
\begin{center}
\includegraphics[width=0.18\linewidth]{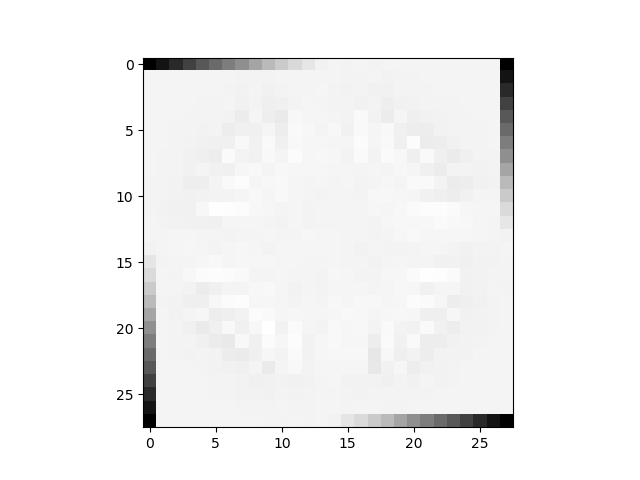}
\includegraphics[width=0.18\linewidth]{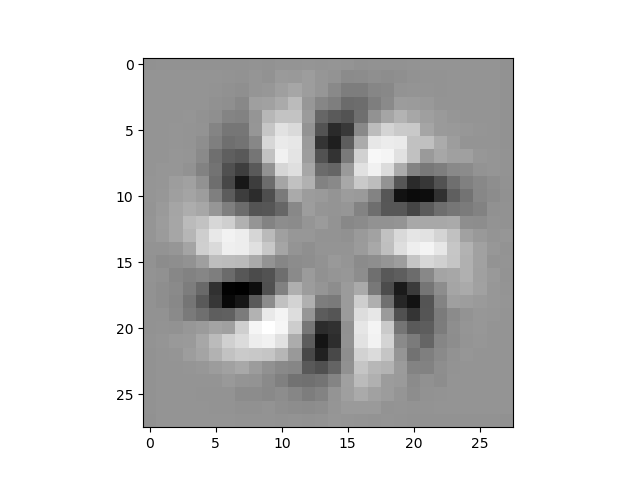}
\includegraphics[width=0.18\linewidth]{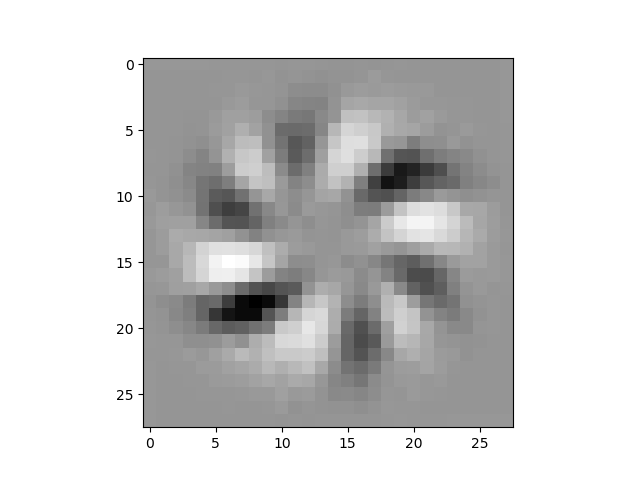}
\includegraphics[width=0.18\linewidth]{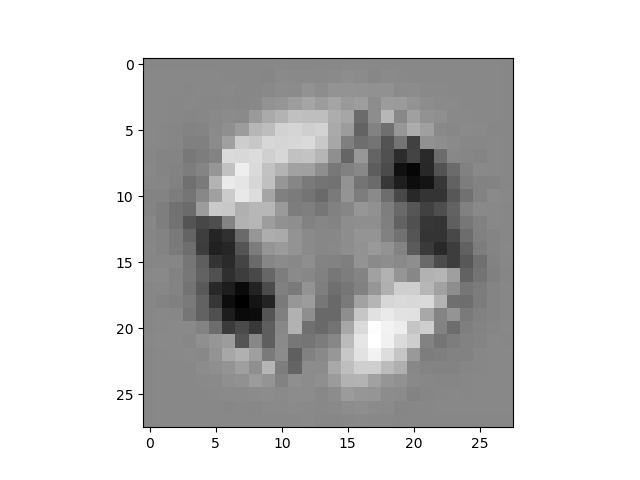}
\includegraphics[width=0.18\linewidth]{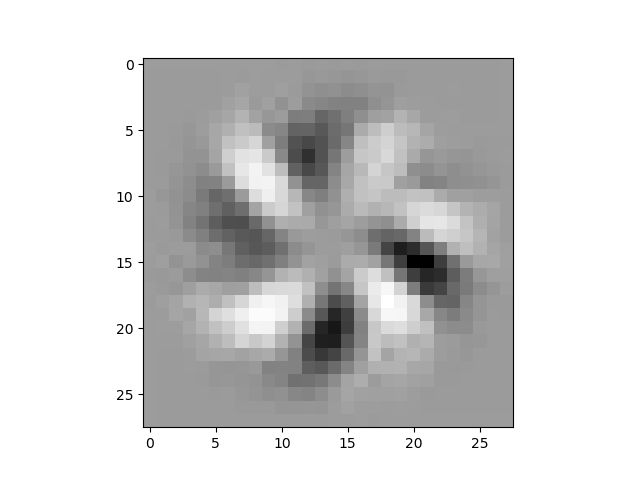}
\end{center}
\caption{On MNIST, Top 5 SVD vectors from image difference}
\label{gcnn-rvh}
\end{figure}

We visualize the top 5 singular vectors in Figure \ref{gcnn-rvh} obtained by the method given in Section \ref{app:invariant} for MNIST data which was intially augmented with random rotations in range $[-180^{\circ}, 180^{\circ}]$. These invariant directions seem to be similar in structure to steerable filters. We believe this underlying structure is useful and of independent interest.

\section{Data sets and model architectures used}
\label{app:arch}
\paragraph{Data sets} MNIST dataset consists of $70,000$ images of $28 \times 28$ size, divided into $10$ classes. $55,000$ used for training, $5,000$ for validation and $10,000$ for testing. CIFAR-10 dataset consists of $60,000$ images of $32 \times 32$ size, divided into $10$ classes. $40,000$ used for training, $10,000$ for validation and $10,000$ for testing. 

All experiments performed on neural network-based models were done using \eat{the validation set of ImageNet and} the test set of CIFAR-10 datasets. 

\paragraph{Model Architectures} 
For the CIFAR-10 based experiments we use the ResNet18 architecture as in \cite{He16}. For the MNIST based experiments we use the StdCNN architecture given in Table \ref{gcnn-table}. This is same as that in \cite{Cohen16}. 

We also evaluated SVD-Universal on rotational equivariant networks. We report our findings with GCNN's\cite{Cohen16} as our base architecture for equivariant networks.  For a fair comparison the GCNN configuration used for CIFAR-10 is the same as that of ResNet18, except that the operations going from one layer to the next are replaced by equivalent GCNN operations,  as given in \cite{Cohen16}.  Similarly the GCNN used for MNIST, has the same configuration as the StdCNN configuration given in Table~\ref{gcnn-table} with equivalent GCNN operations as given in \cite{Cohen16}.  Input training data was augmented with random cropping and random horizontal flips.

{\small
\begin{table}[] 
\caption{Architectures used for the MNIST experiments} \label{gcnn-table}
\begin{center}
\begin{tabular}{ll}
\multicolumn{1}{c}{\bf Standard CNN}  &\multicolumn{1}{c}{\bf GCNN}
\\ \hline \\
Conv(10,3,3) + Relu & P4ConvZ2(10,3,3) + Relu \\
Conv(10,3,3) + Relu & P4ConvP4(10,3,3) + Relu \\
Max Pooling(2,2)  & Group Spatial Max Pooling(2,2)  \\
Conv(20,3,3) + Relu & P4ConvP4(20,3,3) + Relu \\
Conv(20,3,3) + Relu & P4ConvP4(20,3,3) + Relu \\
Max Pooling(2,2)  & Group Spatial Max Pooling(2,2) \\
FC(50) + Relu & FC(50) + Relu \\
Dropout(0.5) & Dropout(0.5) \\
FC(10) + Softmax & FC(10) + Softmax \\
\end{tabular}
\end{center}
\end{table}
}

\section{Conclusion}
We show how to use a small sample of input-dependent adversarial attack directions on test inputs to find a single universal adversarial perturbation that fools state of the art neural network models also works better as networks are made more invariant to a specific transformation. We introduce the study of universal invariant directions, whose visualizations are reminiscent of steerable filters. We believe that a study of how the subspaces of universal adversarial perturbations and universal invariant perturbations change as a model gains invariance can shed more light on invariance vs. robustness trade-off.

\bibliographystyle{splncs04}
\bibliography{notes-universal}

\begin{thebibliography}{10}
\providecommand{\url}[1]{\texttt{#1}}
\providecommand{\urlprefix}{URL }
\providecommand{\doi}[1]{https://doi.org/#1}

\bibitem{Athalye17}
Athalye, A., Engstrom, L., Ilyas, A., Kwok, K.: Synthesizing robust adversarial
  examples. arXiv preprint arXiv:1707.07397  (2017)

\bibitem{Cohen16}
Cohen, T.S., Welling, M.: Group equivariant convolutional networks. In
  Proceedings of the International Conference on Machine Learning (ICML)
  (2016)

\bibitem{Cohen17}
Cohen, T.S., Welling, M.: Steerable {CNN}s. In International Conference on
  Learning Representations  (2017)

\bibitem{Dieleman16}
Dieleman, S., De~Fauw, J., Kavukcuoglu, K.: Exploiting cyclic symmetry in
  convolutional neural networks. In Proceedings of the International Conference
  on Machine Learning (ICML)  (2016)

\bibitem{Dumont18}
Dumont, B., Maggio, S., Montalvo, P.: Robustness of rotation-equivariant
  networks to adversarial perturbations. arXiv preprint arXiv:1802.06627
  (2018)

\bibitem{Esteves18}
Esteves, C., Allen-Blanchette, C., Zhou, X., Daniilidis, K.: Polar transformer
  networks. In International Conference on Learning Representations  (2018)

\bibitem{Gilmer18}
Gilmer, J., Adams, R.P., Goodfellow, I., Andersen, D., Dahl, G.E.: Motivating
  the rules of the game for adversarial example research. arXiv preprint
  arXiv:1807.06732  (2018)

\bibitem{He16}
He, K., Zhang, X., Ren, S., Sun, J.: Deep residual learning for image
  recognition. In Proceedings of the IEEE conference on computer vision and
  pattern recognition pp. 770--778 (2016)

\bibitem{S20invariance}
Kamath, S., Deshpande, A., Subrahmanyam, K.V.: Invariance vs. robustness
  trade-off in neural networks. arXiv preprint arXiv:2002.11318  (2020)

\bibitem{S20universal}
Kamath, S., Deshpande, A., Subrahmanyam, K.V.: Universalization of any
  adversarial attack using very few test examples. arXiv preprint
  arXiv:2005.08632  (2020)

\bibitem{Khrulkov18}
Khrulkov, V., Oseledets, I.: Art of singular vectors and universal adversarial
  perturbations. In: The IEEE Conference on Computer Vision and Pattern
  Recognition (CVPR) (June 2018)

\bibitem{Kondor18Clebsch}
Kondor, R., Lin, Z., Trivedi, S.: Clebsch-gordan nets: a fully fourier space
  spherical convolutional neural network (2018)

\bibitem{Kondor18}
Kondor, R., Trivedi, S.: On the generalization of equivariance and convolution
  in neural networks to the action of compact groups. In Proceedings of the
  International Conference on Machine Learning (ICML)  (2018)

\bibitem{Laptev16}
Laptev, D., Savinov, N., Buhmann, J.M., Pollefeys, M.: {TI}-pooling:
  transformation-invariant pooling for feature learning in convolutional neural
  networks. In Proceedings of the IEEE Conference on Computer Vision and
  Pattern Recognition pp. 289--297 (2016)

\bibitem{Li17}
Li, J., Yang, Z., Liu, H., Cai, D.: Deep rotation equivariant network. arXiv
  preprint arXiv:1705.08623  (2017)

\bibitem{Marcos17}
Marcos, D., Volpi, M., Komodakis, N., Tuia, D.: Rotation equivariant vector
  field networks. In International Conference on Computer Vision  (2017)

\bibitem{Dezfooli17}
Moosavi-Dezfooli, S.M., Fawzi, A., Fawzi, O., Frossard, P.: Universal
  adversarial perturbations. In Proceedings of the IEEE Conference on Computer
  Vision and Pattern Recognition  (2017)

\bibitem{Dezfooli17anal}
Moosavi-Dezfooli, S.M., Fawzi, A., Fawzi, O., Frossard, P., Soatto, S.:
  Analysis of universal adversarial perturbations. arXiv preprint
  arXiv:1705.09554  (2017)

\bibitem{Worrall16}
{Worrall}, D.E., {Garbin}, S.J., {Turmukhambetov}, D., {Brostow}, G.J.:
  Harmonic networks: Deep translation and rotation equivariance. arXiv preprint
  arXiv:1612.04642  (2016)

\end{thebibliography}
\appendix
\section{Fooling rate and error rate.}
\label{sec:defn}
Let ${\cal D}$ be a distribution on pairs of images and labels, with images coming from a set ${\mathcal X} \subseteq {\mathbb R}^d$. In the case of CIFAR-10 images, we can think of ${\mathcal X}$ to be the set of $32 \times 32$ CIFAR-10 images with pixel values from $[0, 1]$. So each image is a vector in a space of dimension 1024.
 Let $(X,Y)$ be a sample from ${\cal D}$ and let $f:{\cal X} \rightarrow [k]$, be a $k$-class classifier.
The natural accuracy $\delta$ of the classifier $f$ is defined to be
\[ \text{Pr}_{(X, Y) \in {\cal D}} [f(X) = Y]. \]
The error rate of a classifier is  $\text{Pr}_{(X, Y) \in {\cal D}} [f(X) \not = Y]$.
 An adversary 
${\cal A}$ is a function ${\cal X} \rightarrow {\mathbb R}^d$. When ${\cal A}$ is a distribution over functions we get a randomized adversary. The norm of the perturbation applied to $X$ is the norm of ${\cal A}(X)$ (we only consider $\ell_2$ norm in this paper).

In Moosavi-Dezfooli et al.~\cite{Dezfooli17,Dezfooli17anal} and Khrulkov and Oseledets~\cite{Khrulkov18}, the authors consider the fooling rate of an adversary. A classifier $f$ is said to be fooled on input $x$ by the perturbation ${\cal A}(x)$ if $f(x + {\cal A}(x)) \neq f(x)$. The fooling rate of the adversary ${\cal A}$ is defined to be
\[ \text{Pr}_{(X,Y)}[f(X + {\cal A}(X)) \neq f(X)].\]
The error rate of ${\cal A}$ on the classifier $f$ is defined to be 
$\text{Pr}_{(X, Y) \in {\cal D}}[ f (X + {\cal A}(X)) \neq Y]$. It is easy to see that 
\begin{equation*}
\begin{split}
& \text{\sf Pr}_{(X, Y) \in {\cal D}}[ f (X + {\cal A}(X)) \neq Y] \\
&= \text{\sf Pr}_{(X,Y)}[ f (X + {\cal A}(X)) \neq Y| f(X) = Y] \ \text{\sf Pr}_{(X,Y)}[f(X) = Y] \\
& + \text{\sf Pr}_{(X,Y)}[ f (X + {\cal A}(X)) \neq Y| f(X) \neq Y] \ \text{\sf Pr}_{(X,Y)}[f(X) \neq Y] \\
& \leq \text{\sf Pr}_{(X,Y)}[ f (X + {\cal A}(X)) \neq Y| f(X) = Y] + (1 - \delta).\\
\end{split}
\end{equation*}
So, if the natural accuracy of the classifier $f$ is high, the fooling rate is close to the error rate.

\section{SVD-Universal on MNIST} \label{svd:mnist}
\begin{figure*}[!h]
\begin{center}
\includegraphics[width=0.49\linewidth]{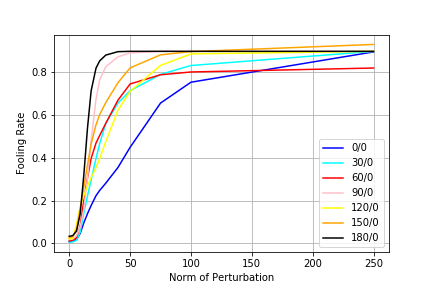}
\includegraphics[width=0.49\linewidth]{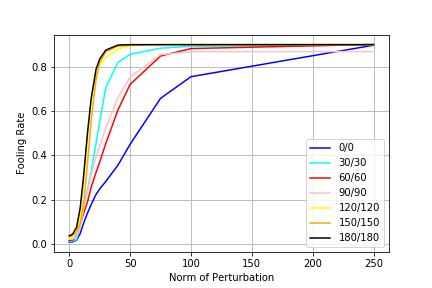}
\end{center}
\caption{On MNIST, error rate of our universal attack using top singular vector of 100 test point gradients, for StdCNN train-augmented with random rotations in range $[-\theta^{\circ}, \theta^{\circ}]$ (left) test unrotated, (right) test-augmented with the same range of rotations as the training set. }
\label{fig:mnist-stdcnn-100-grad-unrot-rot-fool}
\end{figure*}

\begin{figure*}[!h]
\begin{center}
\includegraphics[width=0.49\linewidth]{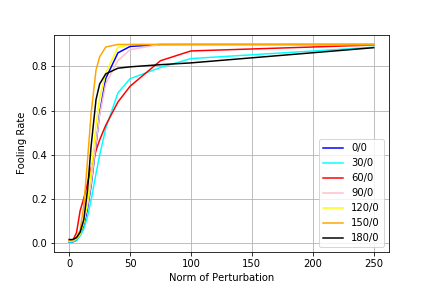}
\includegraphics[width=0.49\linewidth]{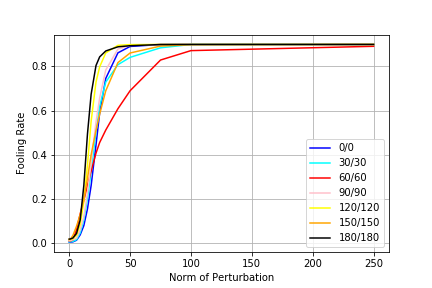}
\end{center}
\caption{On MNIST, error rate of our universal attack using top singular vector of 100 test point gradients, for GCNN train-augmented with random rotations in range $[-\theta^{\circ}, \theta^{\circ}]$ (left) test unrotated, (right) test-augmented with the same range of rotations as the training set. }
\label{fig:mnist-gcnn-100-grad-unrot-rot-fool}
\end{figure*}

Similar to section \ref{sub:large-rot} we evaluate the error rate of SVD-Universal attacks on equivariant networks when the training data is augmented with rotations. We report this on the MNIST data set in Figures~\ref{fig:mnist-stdcnn-100-grad-unrot-rot-fool} and ~\ref{fig:mnist-gcnn-100-grad-unrot-rot-fool} for StdCNN and corresponding GCNN, respectively. We notice similar trend in Figure~\ref{fig:mnist-stdcnn-100-grad-unrot-rot-fool} as in Figure~\ref{fig:cifar10-stdcnn-100-grad-unrot-rot-fool} that the universal attack becomes better as the network achieves better invariance to larger rotations with augmentation.

\end{document}